\documentclass[11pt]{article}
\usepackage{graphicx}
\usepackage{amssymb}
\usepackage{amscd}
\usepackage{mathrsfs}
\usepackage{longtable,lscape}
\usepackage{amsthm}
\usepackage{amsfonts}
\usepackage{amsmath}
\usepackage{bbm}
\usepackage{float}
\usepackage{subfig}
\usepackage{url}
\usepackage{hyperref}
\usepackage{lineno}
\usepackage{csquotes}
\usepackage{wrapfig}
\usepackage[round]{natbib}
\usepackage{caption}
\usepackage[title]{appendix}
\begin{document}

\title{Reply to comments arXiv:2512.07881 and arXiv:2601.06104 on quantum structure in human and AI-generated language} 

\author{Massimiliano Sassoli de Bianchi\footnote{Center Leo Apostel for Interdisciplinary Studies, Vrije Universiteit Brussel (VUB), Pleinlaan 2, 1050 Brussels, Belgium; email addresses: msassoli@vub.ac.be, autoricerca@gmail.com.} $\,$ and  Roberto Leporini\footnote{
        Department of Economics, University of Bergamo, via dei Caniana 2, Bergamo, 24127, Italy; email address: roberto.leporini@unibg.it.}
        }         
\date{}
\maketitle

\begin{abstract} 
\noindent 
We reply to the comments by M.~Sienicki and K.~Sienicki (arXiv:2512.07881) and by K.~Sienicki (arXiv:2601.06104) on our work on quantum-mechanical statistics in human language (arXiv:2407.14924) and on quantum structure in AI-generated language (arXiv:2511.21731). 
We thank the authors for their careful reading and address what we consider to be the main points of criticism: the exploratory nature of the protocol used in the experiments with large language models; the role of marginal-law violations, and of the Contextuality-by-Default criterion, in the identification of entanglement; the limited diagnostic value of a Bose-Einstein fit taken in isolation; the meaning of assigning the lowest energy levels to the most frequent words; and the relation between the vector spaces used by LLMs and quantum state spaces. 
We also correct a typographical error in Table~3 of arXiv:2511.21731, which does not affect the reported CHSH value. 
\end{abstract} 
\medskip

\noindent The comment arXiv:2512.07881 \citep{SienickiSienicki2025} concerns our analysis of Bose-Einstein statistics in human language \citep{Aertsetal2025}, while arXiv:2601.06104 \citep{Sienicki2026a} concerns our study of quantum structure in AI-generated language \citep{Aertsetal2026}, and refers to the former for part of its arguments. K.~Sienicki also privately sent us a longer manuscript \citep{Sienicki2026b}, in which some of these objections are developed further and a few new ones are raised, notably those concerning marginal laws and vector-space representations. Since these are closely related to the points made in the published comments, we address them here as well.
\\

\noindent {\bf Experimental design}. A first criticism, raised in \citet{Sienicki2026a}, is that 
 the experiments with the two LLMs were not conducted under a sufficiently stringent protocol, particularly with regard to the independence of repeated responses and the overall data-collection procedure. We agree that future experiments should employ more rigorous controls, including independent sessions, randomized presentation orders, explicit reset conditions, and a more systematic assessment of run-to-run variability. The aim of the present study, however, was more limited and exploratory. Since the same conceptual-combination tests had already been performed with human participants, our primary objective was to determine whether comparable response patterns, and in particular CHSH violations, would also arise readily in LLMs. The reported results should therefore be understood as a first phenomenological comparison rather than as a definitive experimental characterization of LLM cognition.
\\

\noindent {\bf Marginal laws}. Let us now consider the criticism that, because our data are inconsistently connected and violate the marginal laws, their analysis within the Contextuality-by-Default framework does not reveal contextuality beyond direct contextual influences and therefore does not establish a genuine situation of entanglement comparable to that observed in physics. This objection to our understanding of entanglement in human and artificial cognition is not new. We refer, for instance, to our response in \citep{Aertsetal2018} to the criticisms raised in \citep{dk2014,dkczj2016}, and to the general analysis in \citep{aertsetal2019}, where we examined entanglement in both physical and cognitive systems, including the role of marginal-law violations in its representation within the quantum formalism.

Without entering into the technical details here, our approach understands entanglement primarily as arising from a connection between the entities involved. In the case of conceptual entities, this is a connection of meaning. Correlations are actualized through coincidence measurements on the basis of this connection, and the marginal laws may or may not be satisfied depending on the symmetry and isotropy of the experimental arrangement. From this perspective, the Contextuality-by-Default criterion addresses a precise and important question (whether contextuality remains after direct contextual influences have been taken into account) but it does not exhaust all possible operational interpretations of entanglement. Consequently, subtracting the degree of inconsistent connectedness from the usual CHSH expression should not, in our view, be regarded as a universally decisive test for the presence or absence of entanglement in the broader sense adopted in our work.

It is also worth recalling that theoretical descriptions in physics often rely on idealizations that are only approximately realized in laboratory conditions. Violations of marginal laws have repeatedly been reported in analyses of Bell-type experiments and, in our view, cannot always be dismissed a priori as mere experimental errors; see, for example, \citep{AdenierKhrennikov2007,DeRaedt2012,DeRaedt2013,AdenierKhrennikov2016,Bednorz2017,Kupczynski2017}. The usual assumption that the measurements in a Bell-test scenario are sufficiently separated to admit a single fixed tensor-product representation may therefore fail in actual experimental implementations. One should consequently be cautious about identifying the full phenomenon of entanglement exclusively with the most idealized no-signaling scenario.

A related example concerns photon indistinguishability. For photons to display indistinguishable boson behavior at a beam splitter, their frequencies and arrival times must be sufficiently close; otherwise, they behave as distinguishable entities. See the discussion in \citep{aertsbeltran2020} and the references therein. Thus, even for qualitatively identical photons, indistinguishability is context-dependent. This illustrates more generally why ideal mathematical conditions and their concrete experimental realization should not be conflated.
\\

\noindent {\bf Word shuffling}. As noted in \citet{SienickiSienicki2025}, the words of a text 
 can be arbitrarily shuffled while preserving exactly the same word frequencies. The shuffled text will therefore display the same Bose-Einstein fit as the original one, showing that the fit alone cannot diagnose the presence of narrative meaning. We agree with this observation and have not claimed otherwise. A Bose-Einstein-type frequency distribution is not, by itself, a sufficient condition for semantic organization or understanding. At most, it is one statistical signature that becomes relevant when considered together with independent evidence that a system generates and adapts meaningfully to different semantic contexts. This point was stated explicitly in \citep{SassoliSassoli2026}:

\begin{quote}
It is important to note that the mere existence of a program capable of producing texts that comply with Bose-Einstein
statistics is not, in itself, evidence of semantic understanding, since even a program that simply produces collages
from pre-existing texts would be capable of doing so. The result becomes significant, however, if the program is also
able to adapt dynamically to different semantic contexts. In other words, the claim defended here concerns the
emergence of deeper organizational structures that cannot be reduced to frequency-based regularities alone. It is the
presence of the latter in combination with these deeper organizational structures that confers upon LLMs a distinct
cognitive status. This is why quantum notions such as superposition, contextuality, entanglement-like correlations, and
bosonic amplification mechanisms become explanatorily relevant. At the same time, although both LLMs and quantum
models rely on vector-space formalisms, the precise relation between the semantic spaces of LLMs and the more
richly structured spaces of quantum theory remains an open question for further investigation.
\end{quote}

\noindent Thus, the word-shuffling and similar arguments establish a limitation of the rank-frequency analysis, but it does not render that analysis irrelevant. It shows that the Bose-Einstein fit must be interpreted as one component of a broader evidential framework, rather than as a stand-alone proof of the presence of meaning.
\\

\noindent {\bf Energy ordering}. Both comments question the identification of word energies with ranks \citep{SienickiSienicki2025,Sienicki2026a}. More specifically, Sienicki asks whether 
 the assignment of word energies is merely a rank convention or whether the energies are intended to have semantic or physical content. If the latter is intended, he argues that the energy should ideally be defined by a rule independent of the observed word frequencies, for example through a Hamiltonian.

We agree that the construction of an independently motivated Hamiltonian would be desirable. Zipf's pioneering work already pointed in this direction through his principle of least effort, proposed as an explanation of the power law that bears his name. According to Zipf, speakers and hearers tend to minimize their effort while preserving efficient communication and mutual understanding. This requires a balance between the diversification and unification of words, and therefore an optimization of ambiguity, understood as the capacity of words to express different meanings in different contexts. Zipf also observed that shorter words generally require less physical effort and therefore tend to be used more frequently \citep{Zipf1949}.

Following this line of thought, a Hamiltonian associated with words in a text would plausibly contain at least two contributions. The first would be related to the physical or computational cost of producing a word, and could depend on quantities such as the number of letters, phonemes, syllables, articulatory gestures, or an empirically estimated production time. The second would be a cognitive-semantic contribution associated with the optimization of understanding and with the position of the word within the global semantic organization of the text. Because a word's semantic versatility depends on its relations to the entire semantic environment, this second term would naturally have a mean-field-like or self-consistent character.

These considerations indicate why defining a meaningful Hamiltonian independently of observed word frequencies is a difficult problem. Developing such a construction is an objective for future work and requires a clearer account of how the notion of energy (which historically predates its modern physical formalization) should be extended to cognitive and cultural systems. Nevertheless, irrespective of the precise form that such models may ultimately take, it remains certainly meaningful to assign the lowest energy levels to the most frequently occurring words in a text, and these assignments should not be regarded as a mere convention based on rank.
\\

\noindent {\bf Vector spaces}. We agree with Sienicki that the existence of a vector-space representation does not, by itself, imply a quantum structure. Classical physics also makes extensive use of vectors, without thereby becoming quantum. A vector-space representation becomes specifically relevant to quantum modeling only when the vectors represent states, when measurements and probabilities are defined through an appropriate non-Kolmogorovian structure, and when genuinely nonclassical features such as incompatibility, interference, or entanglement-like correlations are present.

Our claim is therefore not that the real vector spaces used internally by LLMs are automatically Hilbert spaces of quantum theory. Rather, the claim is that the probabilistic and semantic behavior of language can be modeled by quantum-inspired structures defined on vector spaces. This is also the central methodological idea developed in quantum information retrieval \citep{Rijsbergen2004,melucci2015,Aertsetal2018b} and quantum cognition \citep{busemeyerbruza2012}. Of course, the precise mathematical relation between the internal real vector representations learned by LLMs and the more highly structured complex state spaces of quantum theory remains to this day an open problem. 
\\

\noindent {\bf Erratum}. Finally, we take this opportunity to correct a typographical error in Table~3 of \citet{Aertsetal2026}, pointed out in \citet{Sienicki2026b}, which may give the impression of an arithmetic error in one of the reported calculations. 
The calculation itself is correct; the error occurs in the transcription of the data in Table~3. More precisely, it is \emph{Cat Growls} that should have probability $P(A'_2,B_1)=0.222$, whereas \emph{Cat Whinnies} should have probability $P(A'_2,B_2)=0$. With this correction, the expectation value $E(A',B)=0.556$ and the corresponding CHSH value reported in the article remain unchanged.

\end{document}